\documentclass{article}

\usepackage{arxiv}

\usepackage[utf8]{inputenc} 
\usepackage[T1]{fontenc}    
\usepackage{hyperref}       
\usepackage{url}            
\usepackage{booktabs}       
\usepackage{amsfonts}       
\usepackage{nicefrac}       
\usepackage{microtype}      
\usepackage{graphicx}
\usepackage{doi}

\usepackage{graphicx}

\usepackage{tikz}
\usepackage{comment}
\usepackage{amsmath,amssymb} 
\usepackage{color}

\usepackage[accsupp]{axessibility}  

\title{Magnifying Networks for Images with Billions of Pixels}

\author{Neofytos~Dimitriou
~
Ognjen~Arandjelovi\'c\\\
	Department of Computer Science\\
	University of St Andrews\\
    KY16 9SX, United Kingdom\\
	\texttt{\{neofytosd,ognjen.arandjelovic\}@gmail.com} \\
}
\renewcommand{\shorttitle}{\textit{arXiv} Template}
\hypersetup{
pdftitle={A template for the arxiv style},
pdfsubject={q-bio.NC, q-bio.QM},
pdfauthor={David S.~Hippocampus, Elias D.~Striatum},
pdfkeywords={First keyword, Second keyword, More},
}

\begin{document}
\maketitle

\begin{abstract}
	The shift towards end--to--end deep learning has brought unprecedented advances in many areas of computer vision. However, deep neural networks are trained on images with resolutions that rarely exceed $1,000 \times 1,000$ pixels. The growing use of scanners that create images with extremely high resolutions (average can be $100,000 \times 100,000$ pixels) thereby presents novel challenges to the field. Most of the published methods preprocess high--resolution images into a set of smaller patches, imposing an \textit{a priori} belief on the best properties of the extracted patches (magnification, field of view, location, etc.\ ). Herein, we introduce Magnifying Networks (MagNets) as an alternative deep learning solution for gigapixel image analysis that does not rely on a preprocessing stage nor requires the processing of billions of pixels. MagNets can learn to dynamically retrieve any part of a gigapixel image, at any magnification level and field of view, in an end—to—end fashion with minimal ground truth (a single ``global'', slide--level label). Our results on the publicly available Camelyon16 and Camelyon17 datasets corroborate to the effectiveness and efficiency of MagNets and the proposed optimization framework for whole slide image classification. Importantly, MagNets process far less patches from each slide than any of the existing approaches ($10$ to $300$ times less).
\end{abstract}


\section{Introduction}
\begin{figure}[t]
\centering
\includegraphics[width=\linewidth]{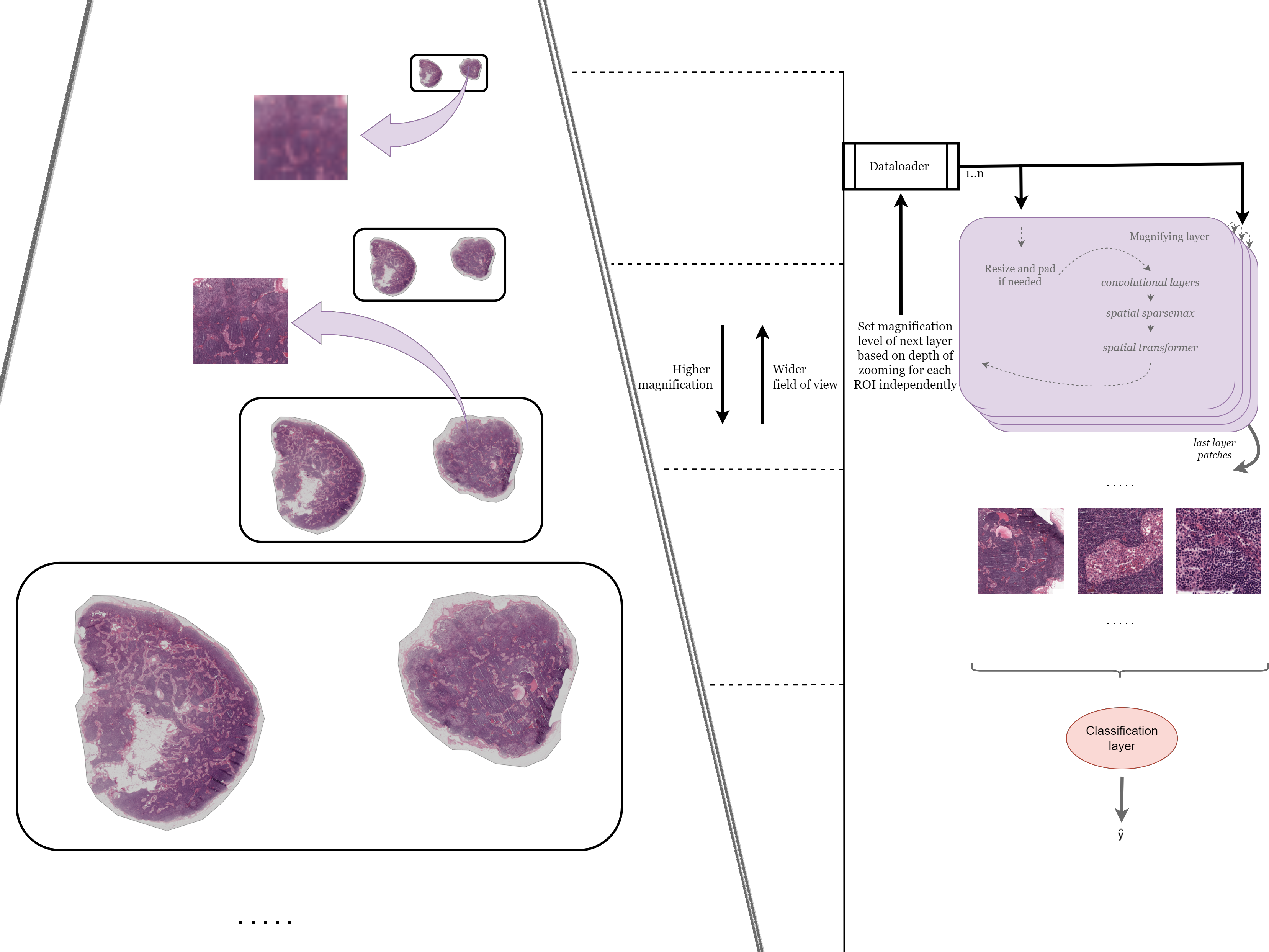}
\caption{An Illustration of the architecture of MagNet. The depicted model consists of four magnifying layers and a classification layer. For each ROI of each magnifying layer, the right level of image resolution is set based on the level of magnification as far. Note that the ROIs of the last layer can span across different magnification levels, and with varying levels of fidelity, thereby providing information across multiple resolutions, and multiple fields of views.}
\label{fig:layout}
\end{figure}

Convolutional neural networks (CNNs) have been the major drive behind the paradigm shift of computer vision towards deep learning. Over the years, a cornucopia of convolution-based network topologies have been described with key differences in their depth, width, and connectivity patterns~\cite{SimoZiss2014,HeZhanRenSun2016,SzegLiuJiaSerm2014,ZagoKomo2016,HuanLiuWein2016,ChenLiXiaoJin+2017,LarsMairShak2016,JadeSimoZissKavu2015,SaboFrosHint2017}. As a consequence of the continuous improvements and the widespread availability of standardized data sets~\cite{DengDongSochLi+2009,KayCarrSimoZhan+2017,KuznRomAlldUijl+2018,EhteVetaJoha+2017,LitjBandEhteGees+2018,AresArauKwokChen+2018,DemiKopeLindPang+2017}, end-to-end trained CNNs dominate the area of visual object recognition.

In parallel with the increase in data availability, hardware advancements have created the possibility of both the capturing and storing of higher resolution images. One of the most extreme yet practically important examples can be found in digital pathology and in particular, in the task of whole slide image (WSI) classification~\cite{DimiAranCaie2019,CaieDimiAran2021}. WSIs are digitized microscope slides, often several gigabytes in size, that have a typical resolution of $100,000\times100,000$ pixels~\cite{YueDimiAran2019,DimiAranCaie2019}.
Conventional neural network optimization using such images is practically infeasible considering the associated memory and compute requirements.

Herein, we introduce a new family of neural networks, henceforth referred to as Magnifying Networks (MagNets), for images that have billions of pixels. MagNets use an attention based mechanism to decide on a course to fine basis the regions of the gigapixel image that need to be analysed, at an increasingly fine scale, in the task and data specific manner of processing extremely large images. Incidentally, this is conceptually similar to a pathologist's knowledge and attention based use of magnification with a brightfield miscroscope. A  microscope  has  multiple  magnification  settings  that enable the user to view a specimen at different scales. Starting  at the  lowest  magnification  setting,  the  entire  specimen can be observed.   As the magnification is increased, finer  detail is accessed, while  at  the  same  time,  a  smaller  part of the specimen is displayed. During  a  visual  examination,  the  clinician  finds  areas  of  interest  at lower magnification levels and then examines them further at higher and higher magnification levels, accruing in the process information from all magnification levels that collectively enable a clinical decision to be made. Similarly, a MagNet starts at the lowest magnification level and recursively identifies, magnifies, and analyses areas of interest with more fine-grain detail (see Figure~\ref{fig:layout}). While remaining within the realm of weakly supervised learning, we extend the spatial transformer module~\cite{JadeSimoZissKavu2015} with a differentiable upsampling mechanism. Depending on the amount of magnification at the current magnifying layer, a version of the WSI at a higher resolution can be accessed by the subsequent layer, as illustrated in Figure~\ref{fig:layout}. MagNets provide a novel way of solving both the ``where'' and ``what'' problems of gigapixel image analysis in an end-to-end fashion~\cite{DimiAranCaie2019}. Importantly, as we show in our experiments, our models can be optimized without the need for extra supervision for the ``where'' problem (e.g.\ boundary boxes). 

In this work, we focus on training and evaluating MagNets for WSI classification~\cite{DimiAranCaie2019}. In particular, we conduct experiments by benchmarking MagNets on the Camelyon data sets. The Camelyon challenge provides a fitting problem for MagNets, considering the varying granularity that needs to be accessed in having to predict macro--metastases, micro--metastasis, and isolated tumour cells (ITC) from WSIs. Given the innate transparency of a model with hard attention, no preprocessing requirements, and ability to perform both localization and classification tasks with no additional information (only slide-level information used~\cite{DimiAranCaie2019}), it would be no exaggeration to say that MagNets have the potential to revolutionize fields, such as Digital Pathology. Our contributions are:
\begin{itemize}
    \item We propose the possibility of identifying and magnifying ROIs starting from a very low resolution downsampled version of the WSI ($56 \times 56 \times 3$ dimensions). We experimentally show that recursively identifying and magnifying ROIs, allows for the extraction of informative areas across magnification levels, without having to preprocess billions of pixels.
    
    \item We introduce a novel form of the spatial transformer module so that we can explore the possibility of ``learning to zoom'' for gigapixel images, without leaving the weakly supervised paradigm. 
    
    \item To the best of our knowledge, this is the first work that automatically finds, and fuses information from multiple \emph{learnt} magnifications on WSIs. The proposed method is able to exploit rich contextual and salient features, overcoming the typical problem of patch--based processing that poorly capture the information that is distributed beyond the patch size. This is an important step towards creating a network architecture that can generalize across different modalities in computational pathology.
    
\end{itemize}

The rest of the paper is organised as follows. In Section~\ref{sec:lit} we provide further context for gigapixel image analysis by discussing the related work in the field, and in particular, the existing ideas for tackling the ``where'' problem. In Section~\ref{sec:magnet}, we introduce the key novelty of the present work, that is, explain the technical underpinnings of MagNets. In Section~\ref{sec:eval}, we provide more details on the data, as well as the optimization framework. Finally, in Section~\ref{sec:discuss}, we discuss our results, and conclude our work with Section~\ref{sec:conclusion}.

\section{Related Work}
\label{sec:lit}
\begin{figure*}[t]
\begin{center}
  \includegraphics[width=1\linewidth]{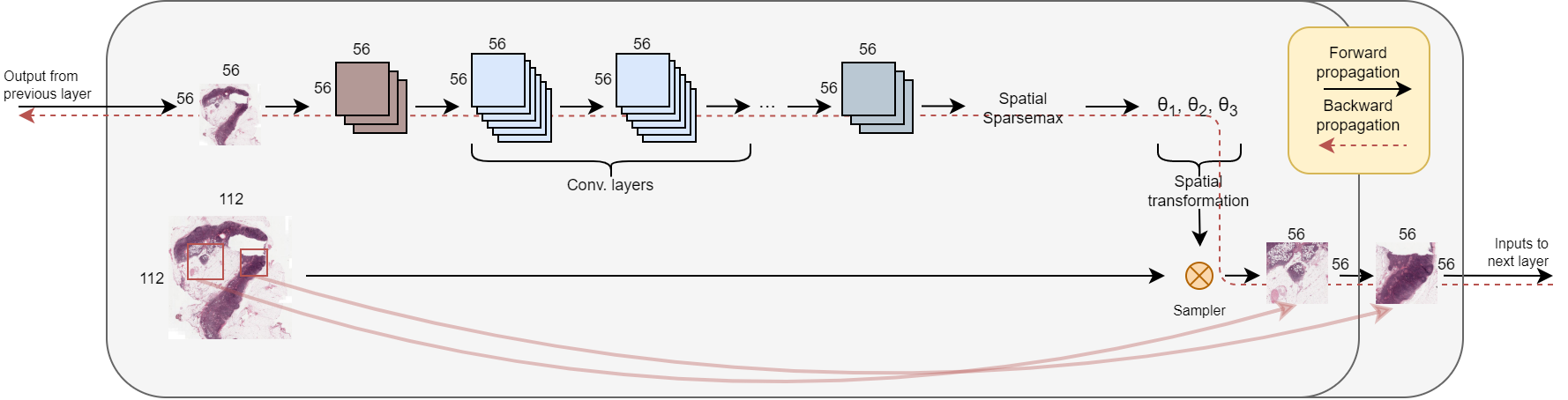}
\end{center}
  \caption{An illustration of a single magnifying layer that outputs two patches. The convolutional layers are independent between the two patches. The red squares illustrate the affine transformation based on the outputted thetas. Note that if this was the last magnifying layer, the image size of the patches would have been $224 \times 224$.}
\label{fig:maglayer}
\end{figure*}
Although other computer vision approaches exist, CNN--based methodologies have emerged as the most effective and popular choice as a way of automatically learning image features rather than handcrafting them~\cite{EhteVetaJoha+2017,AresArauKwokChen+2018}. Therefore, we focus on CNN--based methodologies for WSI classification. Moreover, in order to better highlight the contribution of MagNets, we sort the related works based on their ``where'' problem approach.

A CNN typically excels with image sizes of less than one million pixels~\cite{SimoZiss2014,HeZhanRenSun2016,SzegLiuJiaSerm2014,ZagoKomo2016,HuanLiuWein2016}. Moreover, although they have been a few recent works that explored the use of higher resolution images (e.g.\ up to $8192 \times 8192$~\cite{PincGinnLitj2022}), the current state of the hardware cannot enable CNN--based learning directly from images with billions of pixels. Therefore, on top of optimizing for better visual understanding, practitioners also need to come up with ways for either approximating the spatial distribution of the signal from gigapixel images, or performing some form of dimensionality reduction on the WSIs themselves.

\subsection{Patch extraction}
\subsubsection{Strongly supervised}
One way of identifying and extracting the signal from gigapixel images relies on the use of annotations from domain--specific experts. More specifically, for WSIs, patches based on annotations by pathologist can be extracted in such a way as to ensure a balanced training data set. In essence, the availability of pixel--level annotations turns the ``where'' problem into a trivial challenge that is often addressed with some type of preprocessing pipeline. There exists a relatively large body of work that follows this paradigm~\cite{LiuGadeNoroDahl+2017,WangKhosGargIrsh+2016,LiPing2018,KongWangLiSong+2017,KhenKoriRajkKris+2021,ZhaoYangFangLiu+2020,KoohUnniKhurKris+2021,SuiLiuChenMa+2021}. Most of these approaches extract the ROIs from a single magnification level, e.g. the largest available at $20\times$ or $40\times$. A few, such as the approach of Sui et al.\ ~\cite{SuiLiuChenMa+2021}, instead extract patches by tiling annotated areas at multiple magnification levels. The work of Gecer et al.\ ~\cite{GeceAksoMercShap+2018} is also of particular interest as they built a system conceptually similar to MagNet. Four separate fully CNNs were trained to imitate the actions of pathologists at selecting the right magnification. The training data set was constructed from recordings of pathologists zooming into WSIs to carry out a specific clinical task. The fully--supervised nature of this approach, however, limits its applicability to many clinical tasks for which annotations to this extend is either extremely laborious and expensive, or simply infeasible. 

\subsubsection{Weakly supervised}
In the absence of pixel--level annotations, the literature is divided into three main ways of tackling the ``where'' problem. The most prominent approach is to tile the entirety of a WSI, only perhaps excluding patches that do not meet certain criteria (e.g.\ otsu, entropy, HSV colour space transformation, etc.)~\cite{PiroHeubBerlLadj+2021,LiLiElic2021,HiroYukiAkihRyom2019,LuWillChenChen+2021,DehaCamaMoinLabe+2020,YashAmanLubaChri2021,TellLaakCiom2021,CampSilvFuch2018,HouSamaKurcGao+2016,TellLaakCiom2021}. The second approach involves random sampling from a grid--like patch population~\cite{HashFukuKogaTaka_2020,ChikKimNamGo+2020}. Methodologies that use either of the above two approaches, typically mitigate for the simplistic solution to the ``where'' problem in the later parts of their pipelines. For example, a few recent studies have employed instance--level self--supervision, under the multi--instance learning paradigm, to mitigate for the highly unbalanced nature of tiling (noise--to--signal ratio can be extremely high in WSIs)~\cite{LiLiElic2021,DehaCamaMoinLabe+2020}.

More related to our proposed methodology is the body of work using the third approach. The idea is to allow the right patches to be decided by the model without having to first process them in one way or another. This has been attempted by using different types of attention network~\cite{AichGhas2018,QaisRajp2019,MaksZhaoHobsJenn+2020}. BenTaeib and Hamarneh~\cite{AichGhas2018} employed a recurrent visual attention network to find sub-regions to analyse within $5,000 \times 5,000$ resolution patches. Notably, these high--resolution images were tiled from fixed magnification scales. In fact, processing higher resolution patches came as a consequence of not employing any type of upsampling within the method. On the other hand, Qaiser and Rajpoot~\cite{QaisRajp2019} used a non--differentiable attention network on $1,024 \times 1,024$ resolution patches at $2.5\times$ magnification scale to identify, extract, and process patches from higher, predefined magnification scales ($10\times$ or $20\times$). In general, we find that all of the existing attention--based methodologies on WSI classification still employ heavy preprocessing steps that inevitably impose $a$ $priori$ beliefs on the best magnification scale, field of view, location, etc.\ of the extracted patches, not to mention the computational burden that is associated with them.

The framework we introduce in this paper allows for a multi-resolution, multi-field of view neural network to be optimized in an end-to-end fashion with gigapixel images without requiring initial preprocessing pipelines, and without having to process billions of pixels. 

\subsection{Spatial Transformers}
The most prominent use cases of spatial transformers (STs) are spatial invariance~\cite{ShuChenXieHan2018} and supervised semantic segmentation~\cite{JohnKarpLi2015,DaiHeSun2015,HeGkioDollGirs2017}. To the best of our knowledge, published work on the bridge of WSI classification and STs is nonexistent. Even beyond WSI classification, STs have rarely been trained in a weakly--supervised fashion~\cite{SondSondMaalWint2015,GuoLiuWangLao+2016,AubrKrapBertKlop+2017}. However, the setting of our work differs significantly from these works, and therefore any meaningful comparison cannot be made.

\section{Magnifying Networks}
\label{sec:magnet}
A MagNet consists of $N$ magnifying layers followed by a classification layer. The magnifying layers are responsible for identifying the signal within a WSI, whereas the classification layer is concerned with the visual understanding of the extracted signal in relation to the task at hand. Consider a single gigapixel image $I_0$ that will pass through a MagNet.

\subsection{Magnifying Layer}
\paragraph{Resize and pad.} As we subsequently employ convolutional layers expecting $56 \times 56$ pixel images, input $I$, either as a single input image (e.g.\ $I_0$) or a set of images, is resized to a $56 \times h_i$ or $w_i \times 56$ resolution, based on bilinear interpolation, with the smaller side, $h_i$ or $w_i$, then symmetrically padded (new pixels are black) so that $h_i = 56$ or $w_i = 56$ accordingly. For the purpose of up-sampling (see ``Sampling'' paragraph), a larger version $I^\prime$ ($112 \times 112$ resolution) is also generated using the same protocol. 

Note that although preliminary experiments were conducted using larger images as input to the magnifying layers ($I$ and $I^\prime$ with $112 \times 112$ and $224 \times 224$ resolutions respectively), single GPU training of multiple, stacked magnifying layers was not possible with these resolutions. 

\paragraph{Convolutional layers.} The salient parts in each image (e.g.\ areas with tissue at the lowest magnification level) vary significantly in size. This can be observed in Figure~\ref{fig:filter}. Therefore, the right kernel size for the convolutional operations varies depending on $I$. As such, we stack convolutional operations with different kernel sizes similarly to InceptionNet--v3~\cite{szegedy2015rethinking}. Further information is provided in the Appendix. Our MagNets employ three stacked convolutional layers for each patch independently, e.g.\ a 2-layer MagNet with two patches extracted at each magnifying layer has six of these layers (two at the first layer, and four at the second).

\paragraph{Spatial Transformer.} A STN consists of three parts; a localization network, a grid generator, and a sampler~\cite{JadeSimoZissKavu2015}. 

The \emph{localization network} is typically a FCNN or a recurrent neural network~\cite{SondSondMaalWint2015} that receives an input from a CNN, and its role is to output the parameters of a spatial transformation. With scalability in mind, either options were too prohibitive (GPU VRAM-wise) for MagNet as for each ST there would have been a large amount of parameters that needed to be optimized.

Instead, MagNets utilizes a spatial sparsemax at the last convolutional layer whose output can be used to infer the affine transformation parameters ($s$, $t_x$, $t_y$) directly. In particular, the dimensions of the output of the last convolutional layer are the same as the input image, i.e.\ in our case $56 \times 56$ pixels. Therefore, the output can be thought as a probability mass function, for which following the spatial sparsemax operation, the expected value translates to the scaling parameter ($s$), and expected $L2$ operation over both the x--axis and y--axis corresponds to the translation parameters ($t_x$, $t_y$). More information in provided in the Appendix. 

Given the transformation parameters $s$ for isotropic scaling and $t_x$, $t_y$ for translation on each axis, we further constrain the parameters accordingly:
\begin{align}
s & = max(s, 0.05) \\
t_x& = tanh(t_x) \\
t_y& = tanh(t_y) 
\end{align}
for spatial (affine) transformation $\theta$,
\begin{align}
\theta& =
\begin{bmatrix}
    s & 0 & t_x \\
    0 & s & t_y  \\
\end{bmatrix}
\end{align}

The $tanh$ constrain on the translation parameters implicitly forces the network to favour center extraction, whereas the minimum bound imposed on the scaling helped ensure that we do not get vanishing gradients for some STs during the early stages of training. 

The \emph{grid generator} then creates the desired grid by applying $\theta$ on a meshgrid with dimensions $56 \times 56$. Based on a differentiable \emph{sampler}, e.g.\ billinear sampling, an image can be transformed by $\theta$ by interpolating it onto the grid. 

\paragraph{Sampler} Early on while experimenting with bilinear interpolation, it became obvious that it was a poor choice of a sampler for our work. For more information we direct the reader to the excellent work of Jiang et al.\ ~\cite{WeiWeiwTagl+2019} that later came out corroborating to our empirical analysis and providing an alternative sampler, called Linearized Multi-Sampling, whose gradients are not affected by the amount of scaling (i.e.\ how much the network has zoomed-in) performed (part of our empirical analysis is included in the Appendix). We use the original implementation of this sampler as provided to the authors by Jiang et al.\ ~\cite{WeiWeiwTagl+2019}.

\paragraph{Sampling} This is the part that makes each layer "magnifying", and constitutes our main technical contribution to the field. Given $\theta$ and $I$, we can transform $I$ based on $\theta$ (simple matrix operation) to retrieve an image containing only the detected ROI. MagNet applies the transformation $\theta$ on $I^\prime$ instead, thereby allowing the output to contain information (finer-grain) that was not present in $I$. An example of a magnifying layer that outputs two patches is shown in Figure~\ref{fig:maglayer}. The introduction of new information by transforming images that came from a higher magnification level is what enabled the idea of stacking multiple magnification layers together. 

\subsection{Classification Layer}

The images of the last magnifying layer are sampled using a grid with $224 \times 224$ pixel resolution (instead of $56 \times 56$ pixels). These images are forwarded through an ImageNet--pretrained CNN (InceptionNet-v3) that outputs a feature map into a Gated Recurrent Unit (GRU) network. The output of the GRU is passed through a FCNN (two layers with 512 and 256 hidden neurons respectively) to output a slide--level $\hat y$ estimate on whether the given WSI contains cancer or not.

\subsection{Auxiliary classifiers}
A form of both self--supervision and weak--supervision is introduced by using two auxiliary classifiers. These are ImageNet--pretrained ResNets--18 networks~\cite{HeZhanRenSun2016} that output a slide-level prediction using the extracted images from magnifying layer $1$ and layer $3$ respectively. Cross-entropy is used between the slide-level labels and the ResNet-18 outputs in a weakly--supervised fashion. In addition, a paradoxical loss was also employed as a form of self--supervision~\cite{MaksZhaoHobsJenn+2020}. The premise of the paradoxical loss is that information presented at layer $3$ images should provide an equally good, or better, prediction than that from layer $1$.

\subsection{Configurations}
The network consists of $L$ magnifying layers each of which can access increasingly higher magnification scales as determined dynamically from the degree of zooming (i.e.\ $s$) thus far. At each layer $l$, $P_l$ number of patches are extracted (ROIs). 

A consequence of the recurrent nature of MagNets is that an exponential number of patches are extracted and analysed from a single gigapixel image, if more than 1 patch is extracted per layer. In particular, given a constant $P_l$ across the layers:

\begin{equation*}
  \textrm{Total patches}=
  \begin{cases}
    L, & \text{if $P=1$}.\\
    P^L, & \text{otherwise}.
  \end{cases}
\end{equation*}

We find that a combination of $[2, 3]$ for $P_l$, (i.e.\ 2 ROIs are extracted in some magnifying layers, whereas in others, 3 ROIs are extracted) provides a balance between a sufficient rate of expansion (breadth), while allowing for up to 4-layer MagNets (depth) to be trained on a GPU with $24$ GB of VRAM. 

\section{Evaluation}
\label{sec:eval}
\subsection{Camelyon data sets}
The Camelyon data sets contain WSIs from surgically resected lymph nodes of breast cancer patients. These WSIs were independently curated across multiple hospitals~\cite{EhteVetaJoha+2017,LitjBandEhteGees+2018}. Camelyon16 includes images from 238 normal and 160 cancerous tissue sections whereas the publicly available portion of Camelyon17 has a total of 500 WSIs (318 normal, 182 cancerous). In addition, in the case of metastasis, metadata is available as to the extend of the metastasis (macrometastasis, micrometastasis, or isolated tumour cells (ITC)). Since only a small portion of the data set contains the much more difficult ITC cases, we exclude such cases from the training data set. 

We follow the protocol described in the Camelyon competition website, and in addition set aside $25\%$ of Camelyon17 as a testing set. We shuffle the remaining WSIs from Camelyon17 with the Camelyon16 WSIs, and train on the $80\%$ and validate the better models from the remaining $20\%$. The best MagNets (based on the validation set) are retrained on both the training and validation data, and evaluated on the testing set.

The pixel-level annotations that are available for some of the WSIs of Camelyon are not used in our work. Instead, we only use the binary slide-level label that indicates the presence, or lack thereof, of cancerous cells somewhere in the gigapixel image. 

\begin{figure}[t]
\begin{center}
  \includegraphics[width=1\linewidth]{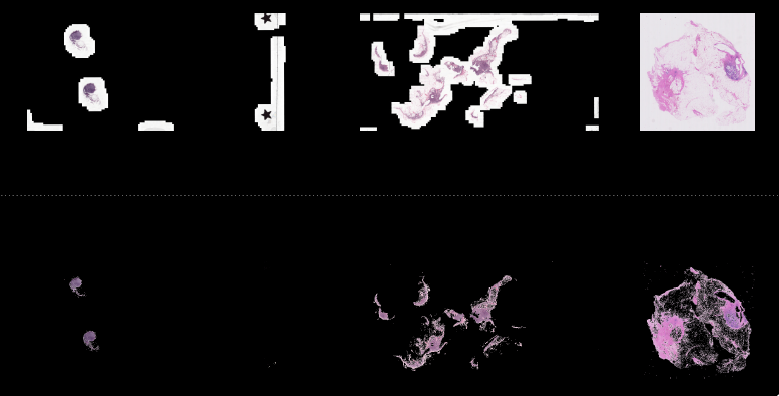}
\end{center}
  \caption{Three WSIs from different hospitals showcasing the differences in the scanning process. The top row shows the unfiltered version of the WSI, whereas the bottom row shows the WSIs after the ``grey'' filter has been applied.
}
\label{fig:filter}
\end{figure}

\begin{figure*}[!t]
\begin{center}
  \includegraphics[width=0.7\linewidth]{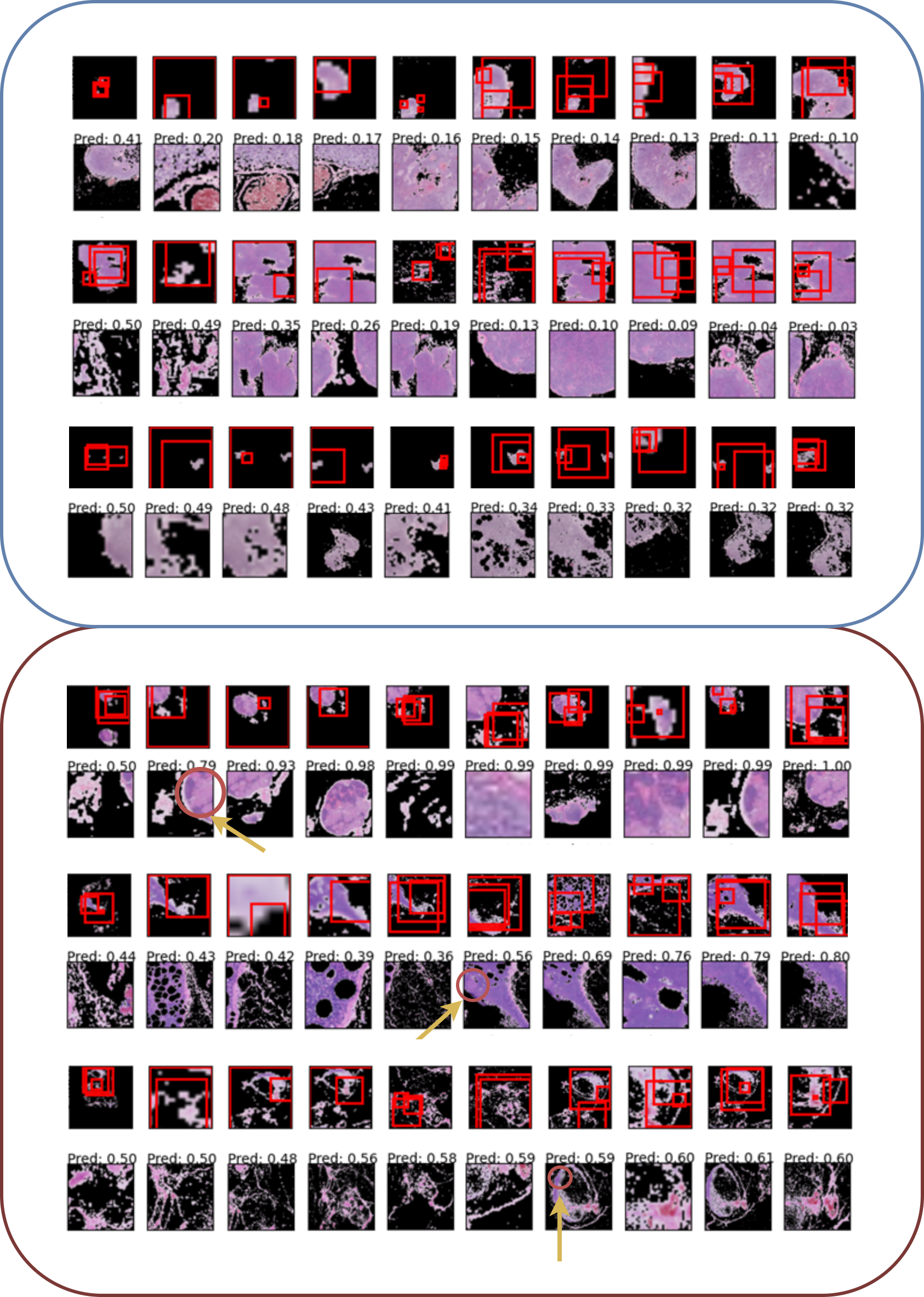}
\end{center}
  \caption{Examples from the testing set against the 3--layer MagNet. WSIs in the top square (blue) do not contain any malignancies, whereas the WSIs on the bottom (red) contain macro-, micro-metastasis, and ITC in order from top to bottom. The pointing arrows, and red circles show the cancer based on the annotations provided by the pathologists. Note that only 10 out of the last layer's 18 patches are shown for space efficiency.
}
\label{fig:layer3runs}
\end{figure*}

\begin{figure*}[!t]
\minipage{0.32\textwidth}
  \includegraphics[width=0.8\linewidth]{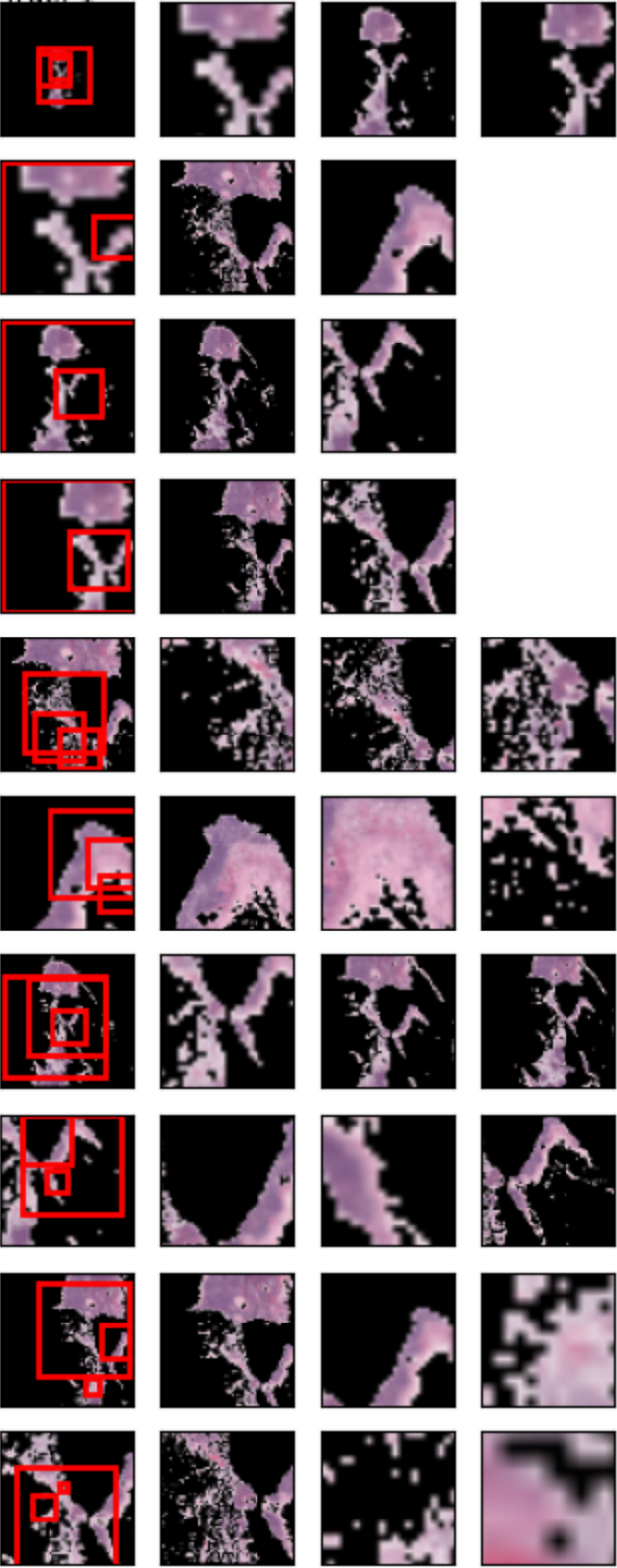}
\endminipage\hfill
\minipage{0.32\textwidth}
  \includegraphics[width=0.8\linewidth]{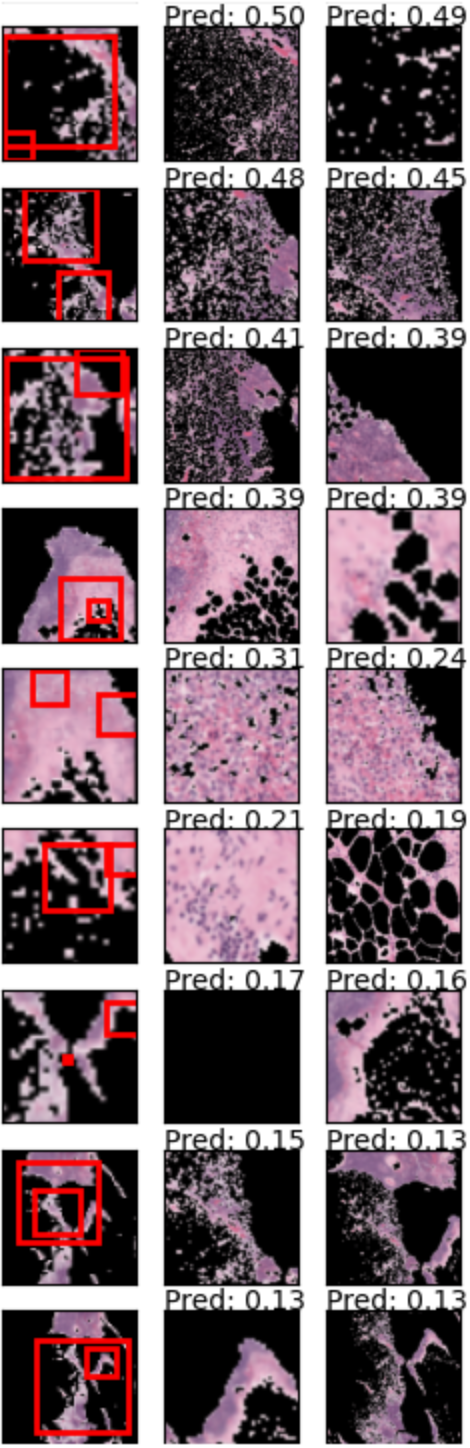}
\endminipage\hfill
\minipage{0.32\textwidth}%
  \includegraphics[width=0.8\linewidth]{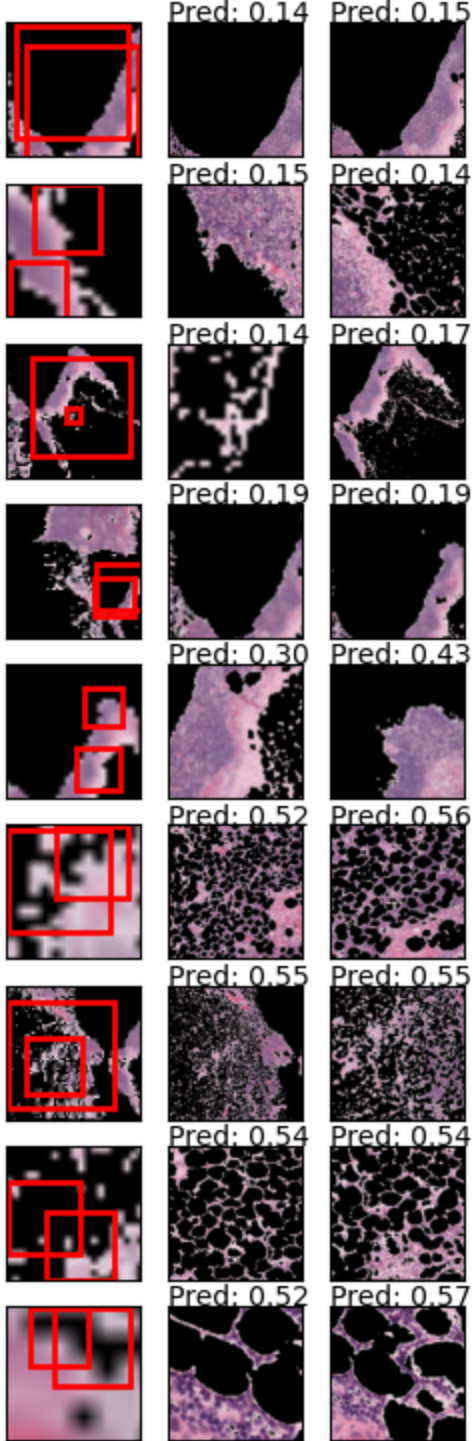}
\endminipage
\caption{A visualization of a forward pass of a WSI with minor metastasis (from the testing set) through a 4-layer MagNet.}
\label{fig:layer4example}
\end{figure*}

\subsection{Data Augmentation}
For any given image, we apply a filter that removes the ``grey'' pixels of the image. This includes any white background, as well as gradients and other artifacts that have red, green, and blue channel values close together (threshold set to $15$). The effects of the filter are illustrated in Figure~\ref{fig:filter}. We employ neither colour normalization nor random colour perturbation~\cite{LiuGadeNoroDahl+2017}. Synthetic data augmentation we perform is based on horizontal and vertical mirroring, and rotation by 90, 180, and 270 degrees.
\subsection{Training}
The networks were trained using the Adam optimizer for 200 epochs. A batch size of $16$ and $8$ was employed for MagNet networks with 3 and 4 layers respectively. The initial learning rate was set to $3 \times 10^-5$ and was decayed using a cosine annealing scheduler. Both ResNet and InceptioNet networks are initialized using pretrained networks on ImageNet. The ST convolutional layers are randomly initialized. 

\paragraph{``Frozen'' patch}
Some WSIs have already been preprocessed so that they only contain regions with tissue, whereas others depict all of the tissue slide (see Figure~\ref{fig:filter}). This diversity comes as a consequence of the differences in the clinical pipelines leading to the creation of WSIs, e.g.\ due to different scanning profiles. In order to mitigate for the above intra--data variance, we freeze the 1st patch of the 2nd layer so that it always attends to the whole input image. This allows for the image to catch-up in quality in the cases where a large amount of zooming was required at the first magnifying layer, i.e.\ when the WSI shows the whole tissue slide.

\paragraph{Loss Functions}
We employ the paradoxical loss function as described by Maksoud et al.~\cite{MaksZhaoHobsJenn+2020}, as a form of self--supervision for the convolutional layers in the weakly--supervised STs. In addition, cross-entropy is used between the slide-level labels and both the last output of the GRU as well as the ResNet-18 outputs.

\begin{table*}
\centering
\caption{The results of MagNets on the testing set containing $25\%$ of the publicly available Camelyon17 data set. }
\begin{tabular}{lccccc}
\hline
                                     & Macro-        & Micro-        & ITC           & Macro- \& Micro- & All           \\ \hline
Mean RGB Baseline                    & 59\%          & 57\%          & -             & 58\%          & -             \\
Baseline w/ patch--level supervision & 91\%          & 63\%          & -             & 77\%          & -            \\
3-layer MagNet                       & \textbf{95\%} & 71\%          & 57\%          & \textbf{84\%} & 71\%          \\
4-layer MagNet                       & 91\%          & \textbf{76\%} & \textbf{63\%} & \textbf{84\%} & \textbf{75\%} \\ \hline
                                     &               &               &               &               &              
\end{tabular}
\label{table:results}
\end{table*}

\section{Results \& Discussion}
\label{sec:discuss}

To evaluate the proposed method, using the optimization framework described in the previous section we trained 3-layer and 4-layer MagNets on the task of cancer detection from WSIs. A summary of the results is presented in Table~\ref{table:results} which shows the area under the receiver operating characteristic curve (AUROC) -- the standard evaluation metric used in the related literature~\cite{WangKhosGargIrsh+2016,TellLaakCiom2021,EhteVetaJoha+2017,LitjBandEhteGees+2018} -- obtained using different configurations of the proposed method. Note that we only present our results since we were unable to identify any existing work on Camelyon17 that employs a more sophisticated than tiling approach to the ``where'' problem in a weakly--supervised fashion.  

Interrogating our findings further, we note that although the 3--layer MagNet achieves better separation between macro-metastasis and normal cases, the 4--layer MagNet does better on micro-metastasis, as well as on the ITC cases. The latter is unsurprising given that a 4-layer MagNet makes use of finer-grain detail (cells are visible in Figure~\ref{fig:layer4example}). Perhaps more importantly, the 4--layer MagNet performs
well on ITC cases with a $63\%$ AUROC despite the lack of ITC examples in our training set.

MagNets exhibit robust and effective exploration capabilities, namely attending to image content in an attention driven manner, exploring slides at varying magnification levels best suited to the task at hand and learning how to fuse relevant information both within the same WSI region and across different regions and magnification levels. In addition, the classifier (in the form of InceptionNet) demonstrates an excellent ability to distinguish normal from cancerous tissue across irrespectively of the magnification scale. Examples corroborating this are shown in Figure~\ref{fig:layer3runs}. It is particularly remarkable to observe cases such as those shown in rows 3--4 of the bottom group of examples in Figure~\ref{fig:layer3runs}, wherein the classifier can be seen to change its decision from ``normal'' to ``cancerous'' when sufficient evidence for cancer is accumulated at any magnification level. These examples were verified by using annotations provided by pathologists.

Figure~\ref{fig:layer4example} shows the forward pass of a single example through a 4-layer MagNet. The Figure is made up of three images. Image A contains $10 \times 4$ sub-images wherein the first column contains the input $I$ to a magnifying layer along red squares indicating the patches to be magnified next, as selected by the MagNet. The rest of the images visualize the last magnifying layer along with its magnified patches of $224\times224$ pixel resolution that are forwarded to the classification layer. Since a GRU is employed in the classification layer, we are able to output a prediction with each new patch, allowing for trivial post-processing analysis. For instance, in the example shown, we observe that the model did not find anything that could be considered as a malignancy for the better part of the last layer's patches. The patches responsible for the change in malignancy probability (from $0.19$ to $0.30$ and then $0.43$) contain a region that was not seen in previous patches and which contains unusual looking tissue. As the final patches are forwarded through the GRU, we observe that the model successfully classifies the WSI as cancerous.

\section{Conclusions}
\label{sec:conclusion}
In this work, we introduced the MagNet -- a neural network consisting of fully-connected, convolutional, and recurrent layers that employ spatial transformers (ST) in a novel manner so as to facilitate attention and data driven recurrent exploration and, ultimately, end-to-end learning over gigapixel images. The built-in hard attention mechanism of MagNets makes them perfectly suited for clinical use. In particular, any machine learning system that is deployed in clinical practice must generate accurate explanations for its decisions that are easily interpretable by a medical expert. The explanations generated by MagNets are visually intuitive for a domain-specific expert to interpret and can be generated on the go without any additional overhead. Moreover, MagNets can be optimized without extra supervision (e.g.\ bounding boxes) for the task at hand. This is of high significance since for most clinical tasks, collecting ground truth data required for a higher degree of supervision is either extremely laborious and expensive, or simply not possible, e.g.\ in the case of patient prognosis.

\bibliographystyle{ieee}
\bibliography{magnet}
\end{document}